# Enhancing $PM_{2.5}$ Data Imputation and Prediction in Air Quality Monitoring Networks Using a KNN-SINDy Hybrid Model

Yohan Choi[1], Boaz Choi[2], Jachin Choi[3]

[1]Katy Taylor High School, Katy, TX

[2]University of Texas, Austin, TX

[3]Washington University, St. Louis, Missouri

## Abstract

Air pollution, particularly particulate matter ($PM_{2.5}$), poses significant risks to public health and the environment, necessitating accurate prediction and continuous monitoring for effective air quality management. However, air quality monitoring (AQM) data often suffer from missing records due to various technical difficulties. This study explores the application of Sparse Identification of Nonlinear Dynamics (SINDy) for imputing missing $PM_{2.5}$ data by predicting, using training data from 2016, and comparing its performance with the established Soft Impute (SI) and K-Nearest Neighbors (KNN) methods. Using $PM_{2.5}$ concentrations from five AirKorea monitoring stations over two years (2016–2017), we evaluated the imputation methods' effectiveness using the Index of Agreement (IOA) for the evaluation period 2017. The KNN-SINDy method consistently outperformed other techniques, maintaining the highest IOA values across all levels of missing data. With 10% missing data, SI achieved an IOA of 0.9575, while KNN achieved 0.9543. The SI-SINDy showed a slight improvement with an IOA of 0.9615, and KNN-SINDy further improved it to 0.9623. With 70% missing data, SI significantly dropped to an IOA of 0.6083, while KNN achieved 0.8534, SI-SINDy 0.7540, and KNN-SINDy 0.8611. These results highlight the added value of integrating the SINDy framework with traditional imputation methods, particularly for handling larger gaps in the data. The findings underscore the potential of physics-informed methods like SINDy to enhance air quality monitoring systems' reliability, contributing to more informed environmental decision-making.

## 1. Introduction

Air pollution, particularly particulate matter ($PM_{2.5}$), poses significant risks to public health and the environment. Accurate prediction and continuous monitoring of $PM_{2.5}$ concentrations are crucial for effective air quality management and policymaking. Due to the irrefutable evidence of causal associations between exposure to ambient pollutants, air quality monitoring (AQM) networks have been established in recent decades to track pollutant levels (Belachsen & Broday, 2022). The higher the spatiotemporal measurement coverage, the greater the ability of regulatory agencies to enforce environmental standards and derive reliable risk estimates, assessing the true cost of air pollution to society. However, monitoring data often suffers from periods of missing records due to device failure, calibration procedures, maintenance, limited communication, or other technical difficulties. Imputation of missing records may be a prerequisite for advanced statistical analyses that require a complete dataset, such as matrix factorization-based source apportionment methods (Shahbazi et al., 2018).

Imputation of AQM stations' data can be carried out using either univariate or multivariate methods. In univariate imputation, each $PM_{2.5}$ time series measured by any AQM station is independently imputed. Univariate imputation can be performed by various methods, such as linear, spline, or Nearest Neighbour (NN) interpolations; spectral methods, e.g., Discrete Fourier Transform (DFT); or variations of Recurrent Neural Networks (RNN), which can capture long-term temporal dependencies (Yuan et al., 2018). In multivariate imputation, the AQM network is viewed as a multivariate dataset, where at each observation time point, several $PM_{2.5}$ measurements are gathered simultaneously at different AQM stations. While univariate imputation relies only on the non-missing data from the imputed series itself (i.e., utilizing temporal relationships), multivariate methods can leverage relationships among different co-measured time series (i.e., accounting also for spatial relationships). Hence, multivariate imputation methods have been found to outperform univariate methods when imputing long periods of missing data (Chaudhry et al., 2019).

Traditional imputation methods, such as mean imputation or linear regression, often fail to capture the underlying temporal and spatial dynamics of $PM_{2.5}$ data (Thongthammachart et al., 2021). Recent advancements in data science have introduced sophisticated techniques like Soft Impute, which leverages low-rank matrix approximation to handle missing data more effectively (Chen & Xiao, 2018). Soft Impute utilizes matrix factorization to fill in missing entries by approximating the original matrix with a lower-rank version, thereby preserving essential patterns and structures within the data. Mazumder et al. (2010) introduced a matrix completion technique with nuclear norm minimization, which enhances the robustness and accuracy of the imputation. While Soft Impute has shown promise in various applications, it still has limitations in capturing the complex, nonlinear relationships inherent in environmental datasets (Pu & Yoo, 2021).

In this study, we explore the application of Sparse Identification of Nonlinear Dynamics (SINDy) for imputing missing $PM_{2.5}$ data. SINDy is a powerful framework that identifies the underlying governing equations of a dynamical system from data, thus offering a more physics-informed approach to imputation (Brunton et al., 2016). By leveraging the sparse nature of the governing equations, SINDy can provide more accurate reconstructions of missing values while maintaining the interpretability of the imputed data. We compared the performance of two multivariate imputation methods: SINDy and Soft Impute. Our dataset comprises $PM_{2.5}$ concentrations from five AirKorea monitoring stations over three years (2016–2018), with hourly observations.

2. **Materials and Methods**

*2.1 Study Area and Data*

We utilized hourly surface $PM_{2.5}$ concentrations observed at five ground-based air quality monitoring stations across Seoul, South Korea (Figure 1) during the period from 2016 to 2018. These observations were sourced from the AirKorea database managed under Korea's Ministry of Environment (Korea Ministry of Environment, 2017; Korea Ministry of Environment, 2018). The data has passed a routine quality assurance procedure. Our dataset comprises, at most, 26,171 hourly observations for each of the five AQM stations whose data were used.

The five stations that operated during the study period are situated in different areas of Seoul, providing comprehensive coverage of the region. The 2016–2018 $PM_{2.5}$ annual means and standard deviations (SD) for these stations were as follows: Station 1: 23.86 (15.76) µg/m³, Station 2: 24.68 (17.87) µg/m³, Station 3: 25.01 (17.37) µg/m³, Station 4: 24.93 (18.67) µg/m³, and Station 5: 24.69 (16.92) µg/m³. These levels exceed the World Health Organization (WHO)

Global air quality guidelines (AQG), recommended $PM_{2.5}$ annual mean of 5 μg/m³ (World Health Organization, 2021).

Our study leverages this comprehensive dataset to compare the performance of two multivariate imputation methods: Sparse Identification of Nonlinear Dynamics (SINDy) and Soft Impute. By evaluating these methods using $PM_{2.5}$ concentration from the five AirKorea monitoring stations, we aim to provide insights into the strengths and limitations of SINDy and Soft Impute in handling missing $PM_{2.5}$ data and predicting future concentrations.

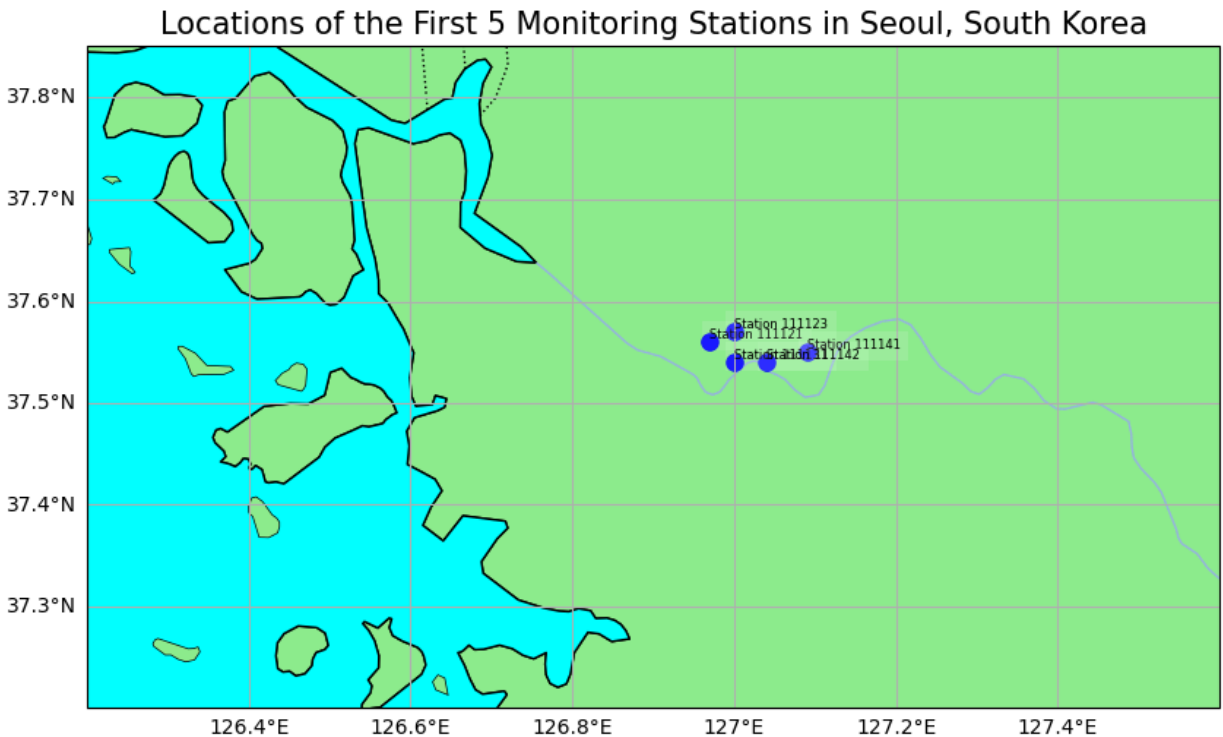

**Figure 1.** Locations of the 5 AirKorea stations whose data were used.

*2.2 Missing-Data Characteristics*

Initial analysis revealed relatively low percentages of missing data across the five stations, with each station having between 2.7% and 3.5% missing values. To provide a robust framework for evaluating the performance of the imputation methods, we introduced additional missing data into the dataset. Specifically, we simulated missing values at random places, ranging from 20% to 60% of the dataset. In addition to random missing values, we also introduced continuous blocks of missing data to mimic realistic operational issues that could arise at monitoring stations. The introduced missing data varied in percentage to assess the robustness of the imputation methods under different conditions. This approach ensures that the evaluation comprehensively covers both sporadic and prolonged periods of missing data.

*2.3 Model description*

In this study, we employed the Sparse Identification of Nonlinear Dynamics (SINDy) framework in combination with imputation methods to address missing $PM_{2.5}$ concentrations from air quality monitoring stations. We aimed to train the SINDy model on data from 2016, which had minimal missing values, and evaluate its performance on data from 2017, which had missing values imputed using two different techniques: Soft Impute and K-Nearest Neighbors (KNN) Impute.

SINDy is designed to uncover the underlying governing equations of a dynamical system from observational data. By leveraging sparsity, SINDy can identify the most relevant features and interactions, providing a model that captures the essential dynamics of the system (Brunton et al., 2016; Champion et al., 2020). In our implementation, SINDy was used to both impute missing $PM_{2.5}$ concentrations and predict future concentrations. The model uses polynomial and differential operators to form a library of candidate functions, from which the most relevant ones are selected through a sparsity-promoting optimization process (Brunton et al., 2016).

Soft Impute is a matrix completion technique that fills in missing data by iteratively performing soft-thresholder singular value decomposition (SVD) (Mazumder et al., 2010). This method is particularly suited for large datasets with substantial missing values, as it maintains the low-rank structure of the data matrix. In our study, the $PM_{2.5}$ concentrations for all the stations for 2016 and 2017 were first normalized to ensure all stations' $PM_{2.5}$ were on the same scale. The Soft Impute method was then applied to fill in the missing values, using an iterative process to approximate the data matrix by a lower-rank version.

KNN Impute is a non-parametric method that estimates missing values based on the values of the nearest neighbors (Troyanskaya et al., 2001; Beretta & Santaniello, 2016). This technique leverages the spatial and temporal correlations in the data, making it suitable for environmental datasets like $PM_{2.5}$ concentrations. The KNN was applied to fill in the missing values by averaging the values of the k-nearest neighbors, with k set to 5 (Beretta & Santaniello, 2016).

After imputation, the data were transformed back to their original scale to ensure consistency with the observed values. These imputed datasets were subsequently used as inputs for the SINDy model to perform predictions. The combination of KNN Impute and SINDy allows the model to benefit from the localized imputation provided by KNN and the dynamic prediction capabilities of SINDy. To systematically assess the methods, missing values were introduced into the 2017 $PM_{2.5}$ data at different percentages (10% to 70%). The missing data were then imputed using both Soft Impute and KNN Impute methods. The SINDy model, trained on the 2016 data, was subsequently used to predict the imputed values. The performance of each method was evaluated using the Index of Agreement (IOA), a robust statistical metric that quantifies predictive accuracy. The IOA was computed for the observed and predicted $PM_{2.5}$ values, allowing for a comprehensive comparison of the imputation methods.

This comprehensive evaluation demonstrated the effectiveness of combining SINDy with traditional imputation methods like Soft Impute and KNN Impute. The KNN-SINDy hybrid model, in particular, showed superior performance, offering a robust solution for accurately imputing and predicting $PM_{2.5}$ concentrations. This enhances the reliability of air quality monitoring data for subsequent analysis and decision-making.

## 3. Results and Discussion

The effectiveness of different imputation methods for handling missing $PM_{2.5}$ concentrations was evaluated using the Index of Agreement (IOA) across varying percentages of missing data. As shown in Figure 1, the SINDy (KNN) Impute method consistently outperforms the other techniques, maintaining the highest IOA values across all levels of missing data. This indicates that the localized imputation provided by the KNN method, when integrated with the dynamic prediction capabilities of SINDy, results in more accurate predictions of $PM_{2.5}$ concentrations. Even with 70% missing data, the SINDy (KNN) Impute method maintains an IOA above 0.85, showcasing its robustness and reliability. Detailed analysis reveals that at 10% missing data, SI achieves an IOA of 0.9575, while KNN achieves 0.9543. The SI-SINDy shows a slight improvement with an IOA of 0.9615, and KNN-SINDy further improves it to 0.9623. At 30% missing data, SI's IOA is 0.9482, KNN's is 0.9450, SI-SINDy is 0.9545, and KNN-SINDy is 0.9571. At 50% missing data, the IOA values are 0.8931 for SI, 0.9253 for KNN, 0.9175 for

SI-SINDy, and 0.9340 for KNN-SINDy. With 60% missing data, SI significantly drops to an IOA of 0.7950, while KNN achieves 0.8991, SI-SINDy 0.8603, and KNN-SINDy 0.9067.

The SI shows a significant decline in performance as the percentage of missing data increases, with the IOA dropping sharply beyond 40% of missing data. This suggests that while SI can handle smaller amounts of missing data effectively, its performance deteriorates significantly with higher percentages of missing data. In contrast, the SI-SINDy method shows improved performance over Soft Impute alone, particularly at higher missing data percentages, though it still falls behind the KNN-SINDy Impute method. The KNN Impute method performs better than SI but is consistently outperformed by the KNN-SINDy hybrid method. This highlights the added value of integrating the SINDy framework with traditional imputation methods, particularly for handling larger gaps in the data. This underscores the importance of selecting appropriate imputation techniques and leveraging advanced dynamic modeling frameworks to improve the quality of air quality monitoring data.

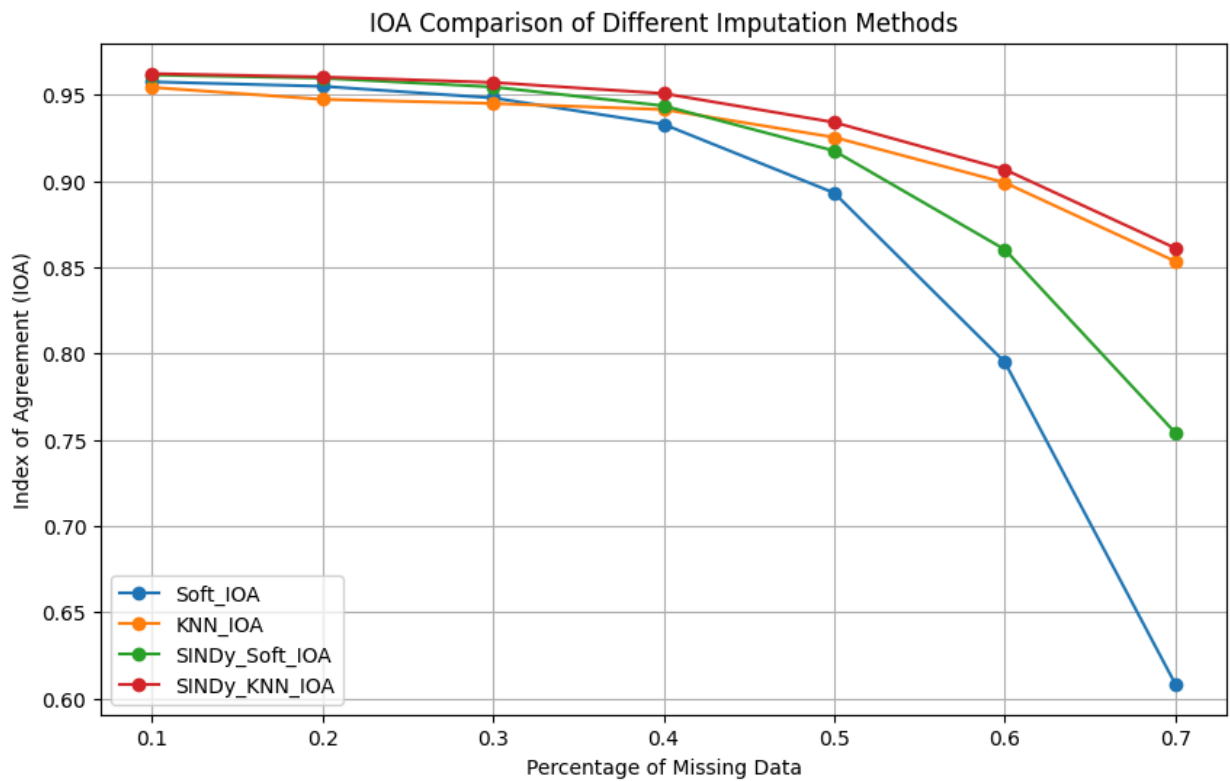

**Figure 2.** It illustrates the Index of Agreement (IOA) for Soft Impute, KNN Impute, SINDy (Soft Impute), and SINDy (KNN) across varying percentages of missing PM2.5 data. The SINDy

(KNN) method consistently maintains the highest IOA, indicating superior performance in handling missing data.

Figure 3 illustrates the actual versus imputed $PM_{2.5}$ concentration values for Station 3 and Station 4, using various imputation methods at varying levels of missing data (10%, 30%, 50%, and 70%). The Index of Agreement (IOA) for each method is also provided to quantify the performance. Although the study included five stations, only the results for Station 3 and Station 4 are presented here for clarity. At 10% missing data, both stations demonstrate that the observations align closely with the imputed values across all methods. For Station 3, SI achieves an IOA of 0.96, KNN Impute 0.96, SI-SINDy 0.97, and KNN-SINDy 0.97. Similarly, for Station 4, the IOA values are 0.96 for SI and KNN Impute, while SI-SINDy and KNN-SINDy both achieve 0.98. This indicates that all methods perform well at this level of missing data, with slight variations in their predictive accuracy, suggesting robustness in the imputation methods, especially when combined with SINDy. Increasing the missing data to 30%, we observe less variability in the IOA values. For Station 3, SI achieves an IOA of 0.95, KNN Impute 0.95, SI-SINDy 0.96, and KNN-SINDy 0.96. Although the performance slightly declines, the KNN-SINDy method continues to exhibit strong alignment with observations, indicating its effectiveness in handling higher missing data percentages. Station 4 shows a similar trend at 30% missing data, with IOA values of 0.95 for SI and KNN Impute, 0.96 for SI-SINDy, and 0.97 for KNN-SINDy. The SINDy methods, particularly KNN-SINDy, maintain a higher accuracy compared to their standalone counterparts.

At 50% missing data, the performance of SI begins to significantly decline with an IOA of 0.91 for Station 3, whereas KNN Impute holds at 0.93. The SINDy-enhanced methods show robust performance with SI-SINDy at 0.92 and KNN-SINDy at 0.94. This indicates that the integration with SINDy helps maintain predictive accuracy even at higher levels of missing data. For Station 4, the trend continues with SI achieving an IOA of 0.89, KNN Impute 0.93, SI-SINDy 0.93, and KNN-SINDy 0.94. The KNN-SINDy method again demonstrates superior performance, highlighting its robustness. At the highest level of missing data (70%), SI's performance deteriorates sharply with an IOA of 0.63 for Station 3. KNN Impute, however, maintains an IOA of 0.85, and the SINDy methods show IOA values of 0.76 for SI-SINDy and 0.87 for KNN-SINDy. This significant drop for SI underscores the advantage of using SINDy, particularly in combination with KNN, for handling extensive missing data. Station 4 at 70% missing data shows SI with an IOA of 0.57, KNN Impute at 0.85, SI-SINDy at 0.77, and KNN-SINDy at 0.87. The plots confirm that while all methods struggle at this level of missing data, the KNN-SINDy method consistently outperforms the others, maintaining a relatively higher accuracy. This demonstrate the superior performance of the KNN-SINDy imputation method across varying levels of missing data. This combination effectively leverages the strengths of localized imputation provided by KNN and the dynamic prediction capabilities of SINDy, making it a robust choice for accurately imputing and predicting $PM_{2.5}$ concentrations in air quality monitoring datasets.

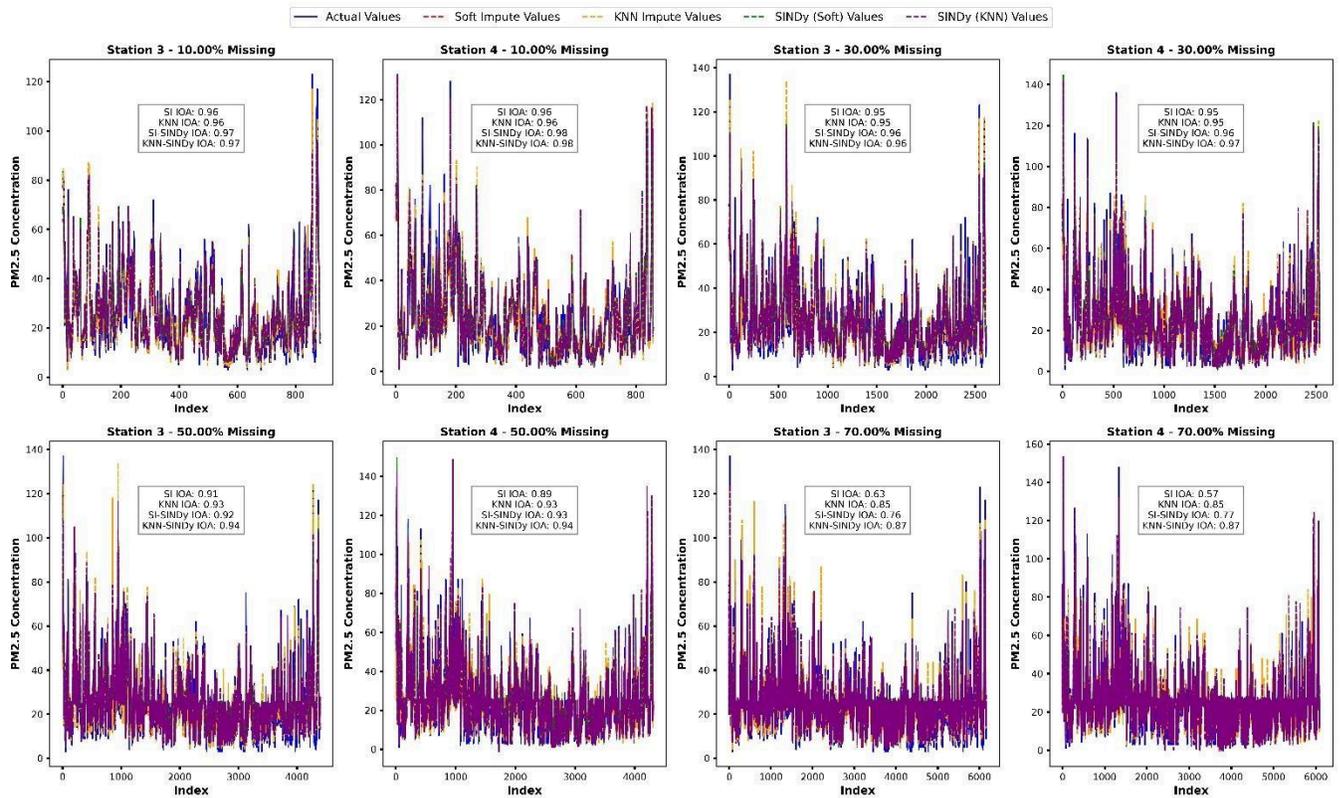

**Figure 3.** Comparison of observed and imputed $PM_{2.5}$ concentration at two monitoring stations (Station 3 and Station 4) with varying percentages of missing data (10%, 30%, 50%, and 70%). The observed $PM_{2.5}$ concentrations is blue, SI as red dashed, KNN impute as orange dashed, SI-SINDy as green dashed, and KNN-SINDy as purple dashed. The inset boxes within each subplot present the IOA for each imputation method, indicating the accuracy of the imputed values relative to the observations.

## 4. Conclusion

This study evaluated the effectiveness of SINDy, Soft Impute (SI), and K-Nearest Neighbors (KNN) methods for imputing missing $PM_{2.5}$ concentrations from air quality monitoring stations. Our comprehensive analysis revealed that the KNN-SINDy hybrid model consistently outperforms the other techniques, maintaining the highest IOA values across all levels of missing data. Detailed analysis at various missing data percentages (10%, 30%, 50%, and 70%) demonstrated that while SI and KNN can handle smaller amounts of missing data effectively, their performance deteriorates significantly with higher percentages. In contrast, the KNN-SINDy method maintains robust performance, showcasing its ability to accurately predict $PM_{2.5}$ concentrations even with extensive missing data. The KNN-SINDy method effectively leverages the strengths of localized imputation provided by KNN and the dynamic prediction capabilities of SINDy, making it a robust choice for handling missing data in environmental datasets and also predicting at the same time. This combination enhances the reliability of air quality monitoring data, ultimately contributing to more informed and effective air quality management and policymaking. Our findings underscore the importance of selecting appropriate imputation techniques and leveraging advanced dynamic modeling frameworks to improve the quality of air quality monitoring systems.